\documentclass[letterpaper, 10 pt, journal, twoside]{IEEEtran}

\PassOptionsToPackage{dvipsnames}{xcolor}

\usepackage{booktabs}

\usepackage{graphicx} %
\usepackage{amsmath}
\usepackage{amssymb}
\usepackage{amsfonts}
\usepackage{bbm}
\usepackage{commath}  %
\usepackage{cuted}  %
\usepackage[hidelinks]{hyperref}
\usepackage[inline]{enumitem}
\usepackage{todonotes}
\usepackage{xcolor}
\usepackage[hang,flushmargin]{footmisc}
\usepackage[export]{adjustbox}  %
\usepackage{tcolorbox} %
\usepackage{pmboxdraw} %
\usepackage{xspace}
\usepackage{flushend}
\usepackage{csquotes}
\usepackage{cite}
\usepackage{threeparttable}
\usepackage{multirow}
\usepackage{makecell}
\usepackage{microtype}

\definecolor{tabblue}{HTML}{1F77B4}
\definecolor{taborange}{HTML}{FF7F0E}
\definecolor{tabgreen}{HTML}{2CA02C}
\definecolor{tabred}{HTML}{D62728}
\definecolor{tabpurple}{HTML}{9467BD}
\definecolor{tabbrown}{HTML}{8C564B}
\definecolor{tabpink}{HTML}{E377C2}
\definecolor{tabgray}{HTML}{7F7F7F}
\definecolor{tabolive}{HTML}{BCBD22}
\definecolor{tabcyan}{HTML}{17BECF}

\usepackage{MnSymbol}

\usepackage{algorithm}
\usepackage[noend]{algpseudocode}

\usepackage{siunitx}

\usepackage{subcaption}
\usepackage{xparse}

\ExplSyntaxOn

\NewDocumentCommand{\panelref}{m}
 {
  \clist_set:Nn \l_tmpa_clist { #1 }

  \int_compare:nNnTF { \clist_count:N \l_tmpa_clist } = { 1 }
   { Panel~\panelref_format:n { #1 } }
   { Panels~\panelref_list:n { #1 } }
 }

\cs_new:Nn \panelref_format:n
 {
  (\subref{#1})
 }

\cs_new:Nn \panelref_list:n
 {
  \seq_set_from_clist:Nn \l_tmpa_seq { #1 }

  \int_case:nnF { \seq_count:N \l_tmpa_seq }
   {
    {2}
     {
      \panelref_format:n { \seq_item:Nn \l_tmpa_seq {1} }
      \ and\
      \panelref_format:n { \seq_item:Nn \l_tmpa_seq {2} }
     }
   }
   {
    \int_step_inline:nnnn {1} {1} { \seq_count:N \l_tmpa_seq }
     {
      \panelref_format:n { \seq_item:Nn \l_tmpa_seq {##1} }

      \int_compare:nNnTF {##1} < { \seq_count:N \l_tmpa_seq - 1 }
       { ,\ }
       {
        \int_compare:nNnTF {##1} = { \seq_count:N \l_tmpa_seq - 1 }
         { , and\ }
         { }
       }
     }
   }
 }

\ExplSyntaxOff
\renewcommand{\secref}[1]{Sec.~\ref{#1}}
\renewcommand{\eqref}[1]{Eq.~(\ref{#1})}
\renewcommand{\figref}[1]{Fig.~\ref{#1}}
\newcommand{\tabref}[1]{Tab.~\ref{#1}}
\renewcommand{\algref}[1]{Algorithm~\ref{#1}}

\newcommand{\etal}{\emph{et al.}}
\newcommand{\para}[1]{\parskip=5pt\noindent\textit{#1}}

\newcommand{\pose}{\boldsymbol{\xi}}
\newcommand{\quaternion}{\boldsymbol{q}}
\newcommand{\ee}{\mathrm{ee}}

\newcommand{\mani}{\mathcal{M}}

\newcommand{\ourmethod}{DynaMAC\xspace}
\newcommand{\ourbench}{DynaBench\xspace}

\title{\LARGE \bf
One Hand Watches The Other: Dynamic Multi-Agent Cooperation for Sample-Efficient Bimanual Manipulation in Dynamic Environments}
\author{
Jan Ole von Hartz, Abhinav Valada, and Joschka Boedecker%
\thanks{Department of Computer Science, University of Freiburg, Germany.}%
\thanks{This work was supported by Carl Zeiss Foundation with the ReScaLe project, and by the BrainLinks-BrainTools center of the University of Freiburg. {Contact: \tt\footnotesize hartzj@cs.uni-freiburg.de}}%
}
\date{July 2026}

\begin{document}
\maketitle

\begin{abstract}
Multi-stream robot manipulation policies achieve unparalleled sample efficiency and generalization by modeling actions relative to environmental reference frames.
However, existing approaches typically assume these frames to be strictly exogenous.
This causal assumption collapses in dynamic settings, such as when a single robot arm manipulates a moving object or when two arms coordinate, where each arm effectively becomes part of the dynamic environment of the other.
We propose \ourmethod{}, a lightweight, policy-agnostic framework that resolves this causal limitation while preserving the sample efficiency, computational speed, and flexibility of multi-stream policies,
\ourmethod{} treats the opposite arm as a dynamic task parameter,  thereby providing a unified formulation for dynamic manipulation and bimanual coordination without requiring an explicit leader-follower relationship.
To rigorously evaluate these capabilities, we introduce \ourbench{}, a novel benchmark for robot manipulation in dynamic environments.
Across both dynamic environments and bimanual manipulation tasks, \ourmethod{} outperforms leading probabilistic and generative baselines by over 35 percentage points \emph{while} requiring 20 times fewer samples.
Crucially, \ourmethod{} generalizes zero-shot from static demonstrations to dynamic environments, substantially simplifying data collection and establishing an elegant bridge toward human-robot collaboration.
\end{abstract}

\section{Introduction}
Real-world robots cannot assume a static world.
They must react to moving objects and coordinate with other agents.
Imitation Learning or Learning from Demonstration, has emerged as a dominant paradigm for robotic manipulation~\cite{chi2023diffusionpolicy}, but it remains constrained by a severe data bottleneck~\cite{von2024art}.
Long task horizons, high-dimensional action spaces, and stringent precision requirements typically demands large amounts of training data.
At the same time, high-quality demonstrations are expensive to collect and particularly scarce in dynamic environments~\cite{fang2026towards}.
Strong inductive biases are therefore essential for reducing data requirements without sacrificing performance.

Multi-stream learning has emerged as an exceptionally sample-efficient solution by embedding the underlying geometric structure of the workspace directly into the policy architecture.
Instead of learning a highly complex direct mapping from raw observations to actions, multi-stream policies decompose the observation-conditioned action distribution into multiple object-centric coordinate frames~\cite{calinon2016tutorial, zeestraten2018programming, von2024art, von2025unreasonable, von2026msg}.
At inference time, the local models, or \emph{streams}, are transformed into a shared world frame and integrated into a joint policy using product-of-experts fusion.
This explicit reliance on 3D geometry yields unparalleled data efficiency~\cite{von2024art, von2026msg}.
Originally developed for Gaussian Mixture Models~\cite{calinon2016tutorial, zeestraten2018programming}, this multi-stream formulation has since been successfully extended to (Mixtures of) Gaussian Processes~\cite{monterolearning, von2025unreasonable} and generative policies such as Flow Matching~\cite{von2026msg}.
When augmented with automated skill segmentation~\cite{von2024art}, per-skill frame selection~skill~\cite{von2024art}, and foundation-model-based pose estimators~\cite{von2024art, wen2024foundationpose, von2025unreasonable}, these policies scale to long-horizon tasks and generalize across novel environments and object instances.
A recent example is MiDiGaP~\cite{von2025unreasonable}, which learns to prepare coffee from a mere five demonstrations and generalizes widely without further training.

\begin{figure}
    \centering
    \includegraphics[width=0.8\linewidth, trim={0 3cm 0 0}, clip]{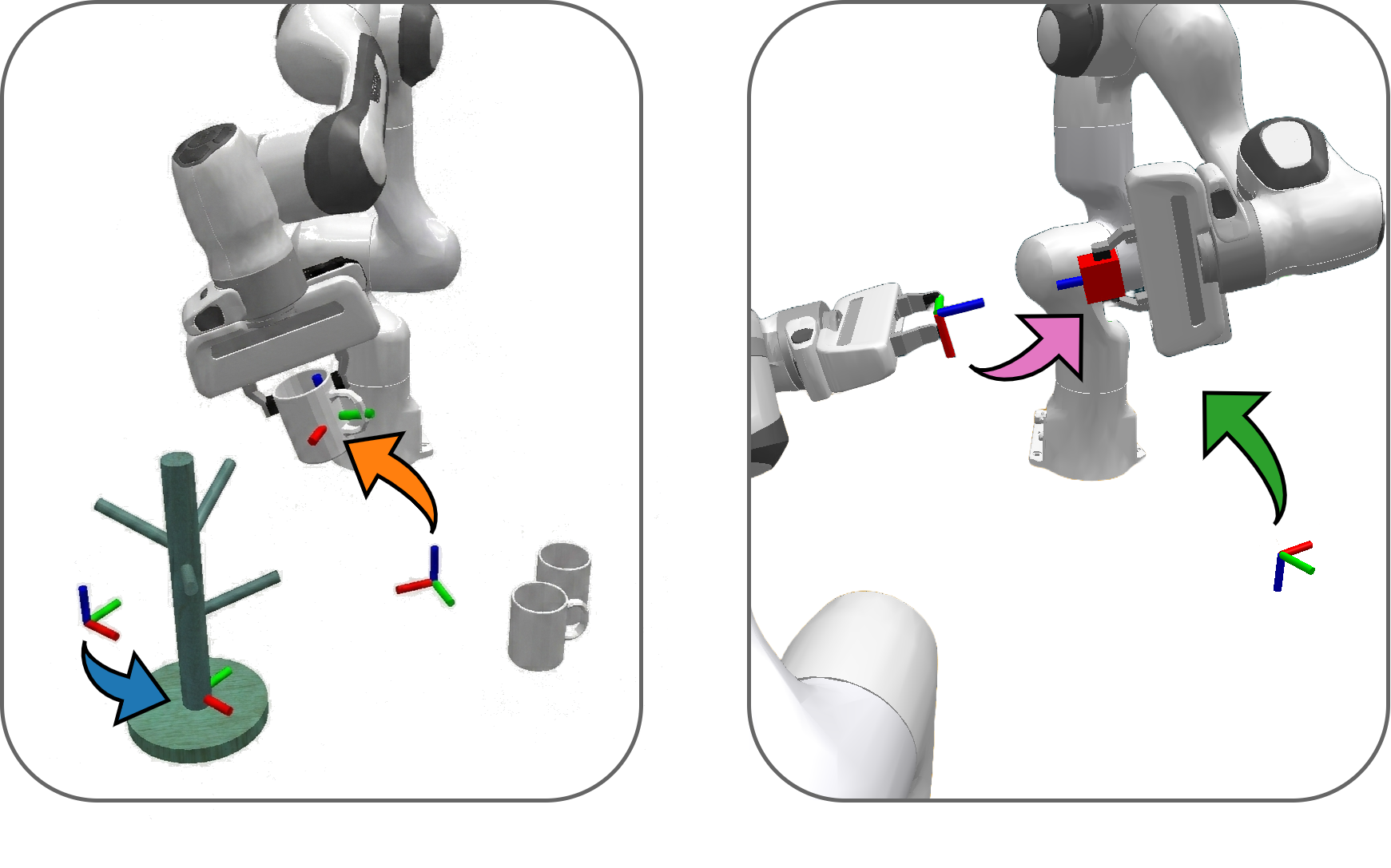}
    \caption{\ourmethod{} brings multi-stream learning to dynamic environments and bimanual manipulation.
    By estimating kinematic links from demonstrations, it distinguishes independently moving objects (\textcolor{tabblue}{blue}) from those manipulated by the robot (\textcolor{taborange}{orange}).
    Disentangling these causal relationships enables multi-stream learning using the dynamic object frames as task parameters.
    These dynamic task parameters further enable effective multi-agent coordination without an explicit leader-follower relationship or coordination module.
    When the cube is lifted (\textcolor{tabgreen}{green}), the right arm's pose informs the handover (\textcolor{tabpink}{pink}).  
    }
    \label{fig:eyecatcher}
    \vspace{-0.3cm}
\end{figure}

Despite these advantages, standard multi-stream learning is fundamentally limited by its reliance on strictly exogenous task parameters.
The reference frames used by the policy are assumed to be determined by the environment and independent of the robot’s own actions.
This assumption breaks down in three important and related classes of manipulation problems: dynamic environments, multi-agent collaboration, and bimanual coordination.
In dynamic environments, such as when picking a workpiece from a moving conveyor belt, the object frame may initially be exogenous but becomes rigidly coupled to the robot once the object is grasped. Continuing to treat this frame as an independent task parameter leads to inconsistent stream fusion and policy failure.
Multi-agent collaboration compounds the problem, since another agent’s actions continuously change the relevant task frames, for example during handovers or cooperative lifting.
Finally, most multi-stream policies are designed for unimanual manipulation, whereas many everyday tasks, such as opening a jar or carrying a large object, require two arms to work in tandem.

In this letter, we address these limitations by introducing \emph{dynamic multi-stream learning}, a unified framework for manipulation in dynamic environments, multi-agent cooperation, and bimanual coordination.
The key idea is to infer kinematic links between the robot and task objects, which we detect directly from the policy streams.
This kinematic analysis then guides the policy stream selection, restoring the causal assumptions of multi-stream learning, as illustrated in \figref{fig:eyecatcher}.
We formalize this solution as \ourmethod{}, a framework that is agnostic to the underlying policy representation. 
For our empirical evaluation across both simulated and real-world tasks, we instantiate \ourmethod{} using MiDiGaP~\cite{von2025unreasonable}, a state-of-the-art few-shot learning method. 
Since DynaMAC operates as a lightweight, policy-agnostic layer, it preserves all the native advantages of the base policy, namely its sample efficiency, wide generalization, interpretability, and inference-time steerability.
Importantly, \ourmethod{} enables policies trained exclusively on static demonstrations to generalize zero-shot to highly dynamic environments, thereby substantially simplifying data collection.
Moreover, it reacts effectively to perturbations of either arm, opening applications in human-robot collaboration, where a human collaborator might act inconsistently, thus requiring dynamic adaptation of the robot.

To systematically evaluate these capabilities, we introduce \ourbench{}, a benchmark for robotic manipulation in dynamic environments built on RLBench~\cite{james2019rlbench}.
\ourbench{} enables controlled evaluation of policies under moving task objects, dynamic reference frames, bimanual coordination, and external perturbations.
Across both simulation and physical robot experiments, we show that \ourmethod{} substantially improves robustness in dynamic and bimanual manipulation tasks while retaining the sample efficiency of the underlying multi-stream policy.\looseness=-1

In summary, our primary contributions are:
\begin{enumerate}
    \item \textbf{\ourmethod{}:} a novel, model-agnostic multi-stream policy framework resolving the causal breakdown of product-of-experts fusion in dynamic environments.
    \item \textbf{Zero-Shot Dynamic Generalization:} We show that \ourmethod{} seamlessly deploys policies trained purely in static settings into dynamic, non-stationary environments without requiring any additional dynamic demonstrations.
    \item \textbf{A Unified Paradigm:} We demonstrate that manipulation in dynamic environments, decentralized multi-agent cooperation, and bimanual manipulation can all be seamlessly governed by this single framework.
    \item \textbf{DynaBench:} We propose and open-source a new benchmark built upon RLBench~\cite{james2019rlbench}, designed to rigorously evaluate robotic manipulation in dynamic environments.
    \item \textbf{Empirical Validation:} We provide extensive experimental validation across both simulation and physical hardware, proving \ourmethod's robustness in complex, dynamic, and bimanual tasks, including perturbations of the robot. 
    \item \textbf{Open Source:} We make our code, models, and benchmark environments publicly available at \url{dynamac.cs.uni-freiburg.de}.
\end{enumerate}

\section{Related Work}
\para{Multi-Stream Policy Learning} composes multiple object-centric policies at inference time.
For an excellent overview, see Calinon~\cite{calinon2016tutorial} and Zeestraten~\cite {zeestraten2018programming} for a Riemannian perspective.
Recently, multi-stream learning has been extended to learning long-horizon tasks~\cite{rozo2020learning, von2024art}, highly-constrained tasks~\cite{von2025unreasonable}, and integrated with RGB-D perception~\cite{von2024art, von2025unreasonable}.
Originally developed for Gaussian policies, multi-stream learning was recently applied to generative policies like Flow Matching~\cite{von2026msg}.
However, prior work focuses on unimanual manipulation and static environments, while bimanual manipulation and dynamic environments remain underexplored.
Silv{\'e}rio \etal{} investigate bimanual manipulation, but only consider static environments - either by fixing a broom to the robot~\cite{silverio2015learning} or by grasping an object simultaneously with both hands~\cite{silverio2018learning}.
In contrast, we develop an approach for dynamic coordination tasks, such as object handover, by unifying multi-arm coordination and acting in dynamic environments as dynamic multi-agent cooperation.

\para{Dynamic Environments} are characterized by objects being moved by forces other than the robot itself, e.g.\ by conveyor belts or other agents such as humans.
The problem was studied in the task planning literature~\cite{schmitt2019modeling}, but less so in Imitation Learning.
Many works on deep learning policies qualitatively show some emergent ability to react to movements of task objects~\cite{vonhartz2023treachery, chi2023diffusionpolicy}, but the problem is seldom studied formally.
In this work, we propose and open-source \ourbench: a new benchmark for manipulation in dynamic scenes, built on RLBench~\cite{james2019rlbench}.
We further extend multi-stream learning to dynamic scenes and unify it with multi-agent cooperation.
Concurrent with our work, Fang \etal{} proposed the DOMINO benchmark for dynamic manipulation~\cite{fang2026towards}, which focuses on anticipating different object dynamics (e.g.\ linear vs.\ abrupt).
In contrast, building \ourbench on RLBench retains the latter's task diversity, as well as compatibility with its other extensions~\cite{grotz2024peract2, pumacay2024colosseum, zheng2022vlmbench, huang2026robostream}.
Additionally, as we detail in \secref{sec:bench}, it maximizes task instance variability.
Future work might integrate DOMINO's advanced dynamic models into \ourbench.

\para{Bimanual Manipulation} often requires tight coordination between both arms, for example, to stabilize an object manipulated by one arm, to hand it over, or to lift it jointly.
Crucially, coordination might be asymmetric, and the \emph{dominant} (or leading) hand might vary across tasks or even change over time~\cite{krebs2022taxonomy}.
While coordination constraints might be modeled explicitly~\cite{sina2016coordinated}, they can also be learned via Imitation Learning.
Existing approaches fall into three broad categories: monolithic policies, leader-follower architectures, and hierarchical approaches.
Monolithic policies~\cite{grotz2024peract2, lee2025interact} jointly predict the actions of both arms, learning bimanual coordination directly from demonstration.
However, such a monolithic approach increases the size of the action space, which tends to reduce sample-efficiency~\cite{zhao2023dualafford}.
ACT therefore employs action-chunking to shrink the action space~\cite{zhao2023learning}.
In contrast, leader-follower (LF) architectures~\cite{zhao2023dualafford, grotz2024peract2} learn two separate policies, where the follower policy is conditioned on the action predicted by the leader policy.
Besides increasing inference latency (as the follower has to wait for the leader's prediction), LF also requires defining a fixed leader a priori.
Depending on the task, the leader arm might need to vary or even change over time in long-horizon tasks.
Finally, hierarchical approaches~\cite{jiang2025rethinking, lu2025anybimanual} learn unimanual policies that are coordinated by a separate module.
In contrast, we propose an architecture that learns bimanual coordination \emph{without} an explicit leader or coordination module.
Instead, coordination is learned from demonstration using common candidate task parameters, and leader-follower roles might change dynamically.
Moreover, existing methods often only predict \emph{sparse} keyposes instead of dense robot actions~\cite{grotz2024peract2,lee2025interact, lu2025anybimanual} -- a limitation in highly-constrained tasks, where robot trajectories must follow precise motion paths~\cite{von2025unreasonable}.
In contrast, our method enables dense action predictions.

\section{Technical Approach}
We present \ourmethod{}: a unified framework for dynamic multi-stream learning and multi-agent cooperation, including bimanual manipulation.
We first briefly recap multi-stream policy learning, focusing on Gaussian policies~\cite{calinon2016tutorial, zeestraten2018programming, von2024art, von2025unreasonable}.
For generative policies, please see~\cite{von2026msg}.
We then tackle the problem of dynamic environments, and finally show how bimanual manipulation can be reduced to dynamic multi-stream learning.

\subsection{Multi-Stream Policy Learning}
Imitation learning seeks a policy \(p(\boldsymbol{a}\mid\boldsymbol{o})\) that maximizes the likelihood of demonstrated actions given a dataset of observation-action trajectories, \(\mathcal{D}=\{(\boldsymbol o_s^n, \boldsymbol a_s^n)_{s=1}^{T_n}\}_{n=1}^N\).
Here, actions \(\boldsymbol{a}\) are absolute end-effector poses \(\pose=\begin{bmatrix}\boldsymbol x & \quaternion\end{bmatrix}^T\) on the manifold \(\mani_{\text{pose}}=\mathbb{R}^3\times\mathcal{S}^3\).
(We disregard gripper actions for brevity.)
Observations \(\boldsymbol{o}\) comprise the current end-effector pose \(\pose_\ee\) alongside a set of reference frame poses \(\{\pose_f\}_{f=1}^F\), called \emph{task-parameters}.
These parameters arise either from ground-truth object poses or from visual estimates~\cite{von2024art, von2025unreasonable}.
We transform the end-effector poses \(\pose_\ee\) into each local coordinate frame \(f\), yielding datasets \(\{\mathcal{D}^{(f)}\}_{f=1}^F\) of local poses \(\pose_\ee^{(f)}\) via
\begin{equation}\label{eq:frametrans}
    \pose_\ee^{(f)} = \begin{bmatrix}
    \quaternion_f^{-1} (\boldsymbol x_\ee  - \boldsymbol x _f) \quaternion_f & 
    \quaternion_f^{-1} \quaternion_\ee
    \end{bmatrix}^T.
\end{equation}
From these datasets, we learn \(F\) independent, object-centric trajectory models \(\{p(\pose_\ee^{(f)})\}_{f=1}^F\), or \emph{streams}.
During inference, the current observation \(\{\boldsymbol x_f, \boldsymbol q_f\}_{f=1}^F\) transforms the streams back into the world frame, yielding global marginals \(\{p(\pose_\ee\,|\, f)\}_{f=1}^F\).
For a Riemannian Gaussian with parameters \(\{\boldsymbol\mu_k, \boldsymbol\Sigma_k\}\), and \(\boldsymbol A_f\) as the rotation matrix for \(\boldsymbol q_f\), this transformation is given by\looseness=-1
\begin{equation}
\label{eq:gaussiantrans}
\begin{aligned}
    \hat{\boldsymbol\mu}_k
    &= \operatorname{Exp}_{\boldsymbol x_f}
       \!\left(
       \boldsymbol A_f
       \operatorname{Log}_{\boldsymbol e}
       (\boldsymbol\mu_k)
       \right),\\
    \hat{\boldsymbol\Sigma}_k
    &= \left(
       \boldsymbol A_f
       \boldsymbol\Sigma_k
       \boldsymbol A_f^T
       \right)_{\smash{\raise0.3ex\hbox{$\Vert_{\hat{\boldsymbol x}_f}^{\hat{\boldsymbol\mu}_k^f}$}}}.
\end{aligned}
\end{equation}
Assuming conditional independence, we fuse them into a single policy as a product-of-experts via the product of Gaussians, i.e.\
\begin{equation}\label{eq:comb}
   p\left(\pose_\ee\mid \{f\}_{f=1}^F\right) \propto \prod_{f=1}^F p(\pose_\ee\mid f).
\end{equation}

\begin{figure}[t]
    \centering
    \includegraphics[trim={0 0 0 .67cm},clip]{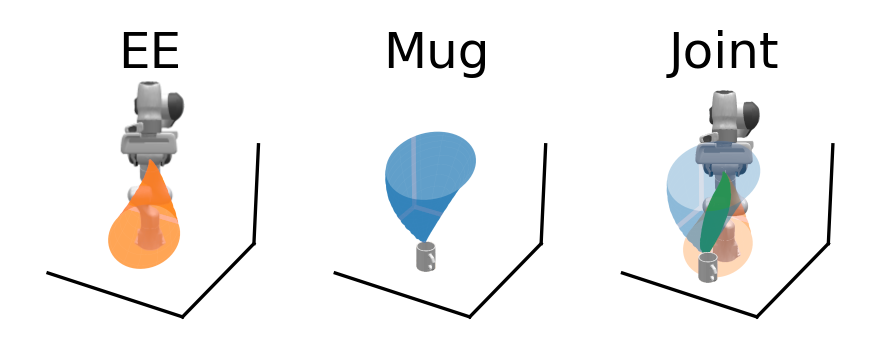}
    \vspace{-0.3cm}
    \caption{Multi-Stream learning models the policy in multiple local coordinate frames, here: the static end-effector frame (\textcolor{taborange}{orange}) and the mug frame (\textcolor{tabblue}{blue}).
    At inference time, these streams are composed as a product-of-experts (\textcolor{tabgreen}{green}).}
    \label{fig:multistream}
    \vspace{-0.3cm}
\end{figure}

\figref{fig:multistream} summarizes this process.
The product-of-experts in \eqref{eq:comb} weights the streams by their precision.
As streams tied to irrelevant background clutter exhibit low precision, we can filter them out~\cite{von2024art}.
In our experiments, we use MiDiGaP~\cite{von2025unreasonable}, a state-of-the-art Gaussian policy learning method, and pair it with TAPAS~\cite{von2024art} to automatically segment skills and discard irrelevant task parameters based on their relative precision.

\subsection{Manipulation in Dynamic Environments}\label{sec:man_dyn_env}
In principle, multi-stream learning accommodates dynamic environments.
Since the pose \([\boldsymbol x_{f,t}\ \boldsymbol q_{f,t}]^T\) of a task-parameters \(f\) might vary with time \(t\), the formation of the local poses \(\pose_{\ee,t}^{(f)}\) in \eqref{eq:frametrans} becomes time-dependent~\cite{calinon2016tutorial}:
\begin{equation}\label{eq:frametrans_time}
\pose_{\ee,t}^{(f)} = \begin{bmatrix}
\quaternion_{f,t}^{-1} (\boldsymbol x_{\ee,t} - \boldsymbol x_{f,t}) \quaternion_{f,t} &
\quaternion_{f,t}^{-1} \quaternion_{\ee,t}
\end{bmatrix}^T.
\end{equation}
During inference, the current task-parameter poses transform the marginals in \eqref{eq:gaussiantrans} before the streams are fused.

\begin{figure}[t]
    \centering
    \includegraphics[width=0.8\linewidth]{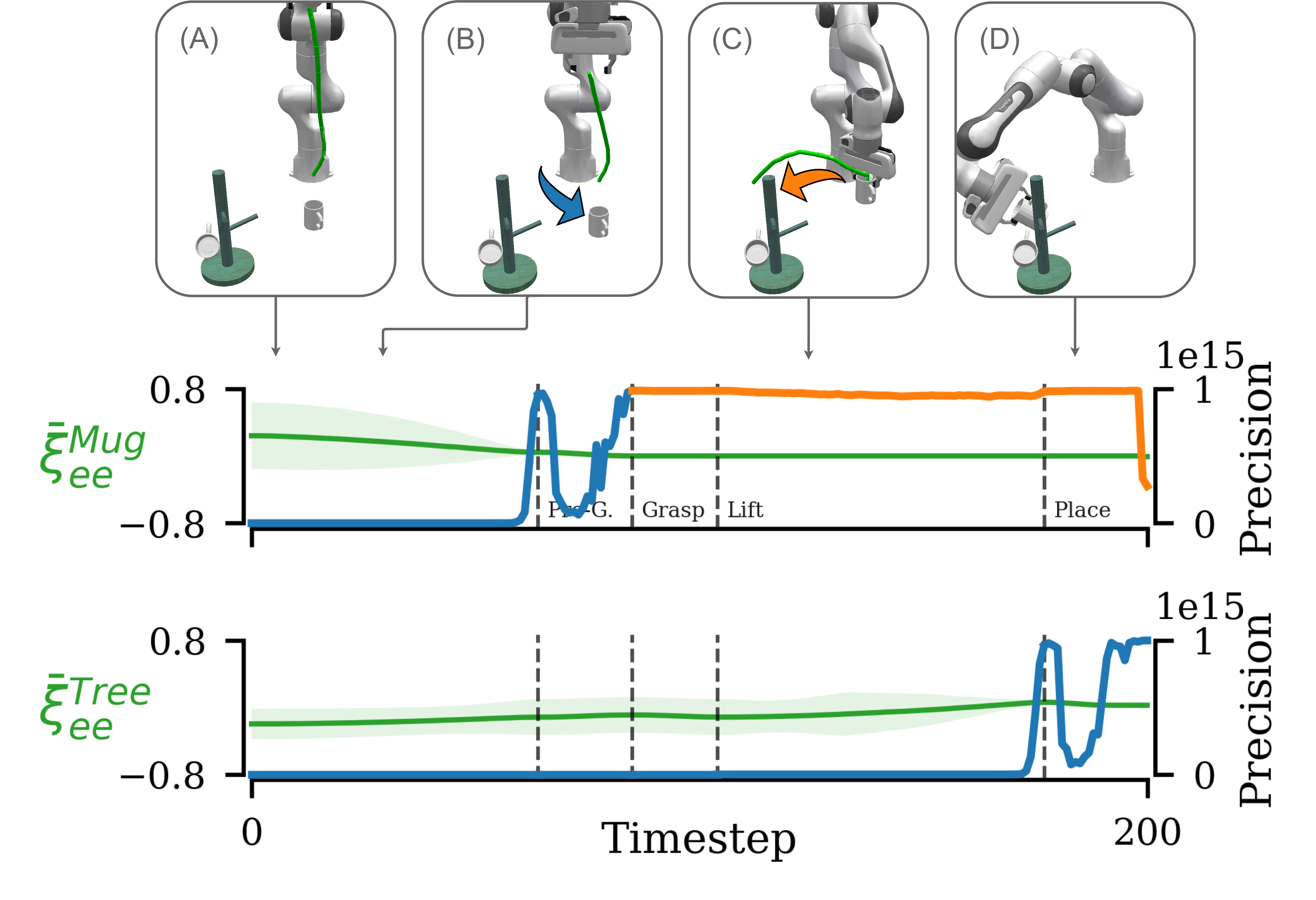}   
    \caption{Kinematic link detection.
    The left y-axes show the local end-effector poses (\textcolor{tabgreen}{green}), averaged over the pose dimensions; the shaded region indicates the total standard deviation.
    The right y-axes show the precision determinants, where \textcolor{tabblue}{blue} denotes no kinematic link between object and robot, and \textcolor{taborange}{orange} denotes a link.
    Dotted lines indicate skill boundaries estimated with TAPAS~\cite{von2024art}. 
    \panelref{subfig:begin}: initially, neither mug nor tree is linked to the robot. 
    \panelref{subfig:move}: external mug motion (\textcolor{tabblue}{$\rcurvearrowright$}) triggers an updated prediction.
    \panelref{subfig:grasped}: after grasping, the mug becomes kinematically linked and follows the end-effector (\textcolor{taborange}{$\rcurvearrowright$}).
    \panelref{subfig:placed}: the link ends after placement.
    Short-lived precision peaks at pre-grasp and during mug–tree alignment do not constitute kinematic links.
    }\label{fig:kinematic_analysis}
    \phantomsubcaption\label{subfig:begin}%
    \phantomsubcaption\label{subfig:move}%
    \phantomsubcaption\label{subfig:grasped}%
    \phantomsubcaption\label{subfig:placed}%
    \vspace{-0.3cm}
\end{figure}

\figref{fig:kinematic_analysis} highlights a causal assumption that undermines standard multi-stream learning in dynamic scenes.
Each stream $p(\pose_\ee \mid f)$ treats the reference frame $f$ as an exogenous variable, modeling a unidirectional causal effect $f \rightarrow \pose_\ee$.
I.e.\, the framework assumes that this conditional density represents an interventional distribution $p(\pose_\ee \mid \mathrm{do}(f))$.
This assumption collapses the moment the robot manipulates the object corresponding to \(f\).
Consider the manipulation sequence in \figref{fig:kinematic_analysis}.
Before the robot grasps the mug (\panelref{subfig:begin, subfig:move}), causal influence flows strictly from the mug to the robot ($\mathrm{Mug} \rightarrow \pose_\ee$), guiding the end-effector toward the mug.
Once grasped (\panelref{subfig:grasped}), the mug's pose is rigidly dictated by the robot, reversing the causal path ($\pose_\ee \rightarrow \mathrm{Mug}$).
During the subsequent co-movement, the conditional density $p(\pose_\ee \mid \mathrm{Mug})$ exhibits near-zero variance.
As the product-of-experts in \eqref{eq:comb} weights the streams by their precision, this single stream then dominates the joint model.
Operating under the flawed assumption that the mug moves independently, the policy fails to initiate motion, collapsing instead to a constant prediction.
However, simply excluding dynamic task parameters would disallow the robot from operating in changing environments altogether.\looseness=-1

We resolve this conundrum by dynamically filtering out frames the moment they become kinematically linked to the end-effector, thereby restoring the exogeneity of the task parameters.
In this way, we disentangle the competing causal directions $f \rightarrow \pose_\ee$ and $\pose_\ee \rightarrow f$.
As outlined in \algref{alg:dynamic_multistream}, we proceed in four steps.
First, we segment long-horizon demonstrations into shorter skills using TAPAS
\figref{fig:kinematic_analysis} shows how the \texttt{PlaceMug} task is segmented into shorter skills, such as grasping and placing.
Within each skill, we identify kinematic links using the stream precisions.
Since TAPAS already pre-computes these precisions, this introduces minimal computational overhead.
Third, we supplement virtual end-effector frames at each skill boundary to compensate for the removed dynamic object frames and to provide additional spatial reference.
Finally, we apply TAPAS to select the relevant task parameter for each skill.

\begin{figure}[t]
    \centering
    \includegraphics[width=0.7\linewidth, trim={0 0 0 0.25cm},
        clip]{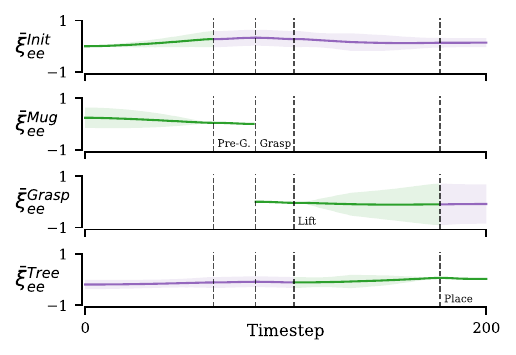}
    \caption{Virtual end-effector frames supplement the dynamic object frames.
    \textcolor{tabgreen}{\textit{Green}} frames are selected, whereas \textcolor{tabpurple}{\textit{purple}} frames are not.
    After grasping, the mug is kinematically linked with the end-effector and \ourmethod{} masks it out to prevent model collapse.
    Instead, a virtual Grasp frame is created to drive post-grasp behavior.
    Likewise, an Init frame drives pre-grasp behavior.}
    \label{fig:virtual_frames}
    \vspace{-0.3cm}
\end{figure}

\begin{figure}[t]
    \centering
    \includegraphics[width=0.8\linewidth,trim={0 .3cm 0 0},clip]{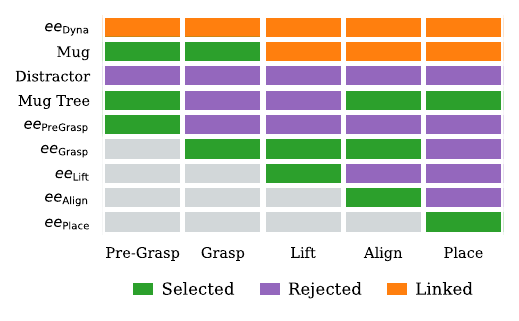}
    \caption{Frame selection result for \texttt{PlaceCups}.
    The Mug frame is used initially, but is ignored once kinematically linked.
    The virtual end-effector frames complement the dynamic frames.
    For example, \(\mathit{ee}_\mathrm{Grasp}\) guides grasping and lifting, while  \(\mathit{ee}_\mathrm{Place}\) guides placing.
    Distractor frames are rejected entirely.}
    \label{fig:frame_selection}
    \vspace{-0.3cm}
\end{figure}

\textit{Kinematic Link Analysis:} As \figref{fig:kinematic_analysis} illustrates, when the robot grasps an object, the corresponding stream's variance drops to near zero, reflecting high mutual information between the end-effector and the object pose.
Let \(\boldsymbol{\Lambda}_{t}^{(f)} = (\boldsymbol{\Sigma}_{t}^{(f)})^{-1}\) be the precision matrix of stream \(f\) at time \(t\).
While sustained spikes in its determinant  \(\det(\boldsymbol{\Lambda}_{t}^{(f)})\) indicate a kinematic link, this raw magnitude is highly sensitive to regularization and difficult to threshold intuitively.
Instead, we identify links using the Geometric Mean Standard Deviation:
\begin{equation}\label{eq:gmsd}
    M_t^{(f)} = \left\lvert \det\left(\boldsymbol{\Lambda}_{t}^{(f)}\right)\right\rvert^{-\frac{1}{2d}},
\end{equation}
where \(d=6\) is the state dimension.
\(M\) measures the average spread across all spatial dimensions, enabling intuitive thresholding.
If \(M_t^{(f)} < \tau_M\), the end-effector and frame \(f\) are kinematically linked at time \(t\).
For instance, a threshold of \(\tau_M=0.001\) corresponds to a mean deviation of \(\SI{1}{\mm}\) or \(\SI{0.001}{\radian}\).
Alternatively, the weighted precision \(\boldsymbol{W}\boldsymbol{\Lambda}_t^{(f)}\boldsymbol{W}\) with diagonal weight matrix \(\boldsymbol{W}\) offers more fine-grained control over position and rotation.
In our experiments, kinematic links were sufficiently apparent even without weighting.%
\footnote{Highly precise pre-grasps can occasionally cause brief, false-positive precision spikes (\figref{fig:kinematic_analysis}). These artifacts can be safely ignored or filtered via temporal variance. Because this phase of the trajectory is fully determined by the pre-grasp pose, temporarily dropping the object frame introduces no spatial ambiguity.
In our experiments, we found both strategies to be effective.}

Our kinematic link analysis and TAPAS' task-parameter selection are grounded in Information Theory, a theoretical connection that has thus far been missing from the literature.
For Gaussian policies, the conditional entropy $H(\pose_\ee \ |\ f)$ is proportional to $\log\det(\boldsymbol{\Sigma}_{t}^{(f)})$. 
Consequently, the mutual information $I(\pose_\ee ; f) = H(\pose_\ee) - H(\pose_\ee \mid f)$ scales proportionally to $-\log\det(\boldsymbol{\Sigma}_{t}^{(f)})$. 
From this perspective, TAPAS selects parameters that maximize $I(\pose_\ee ; f)$, and \ourmethod{} isolates the pathological case where a kinematic link drives spatial variance to zero, causing $I(\pose_\ee ; f)$ to spike. 
This formulation formally bridges our multi-stream approach with recent work in task graph generation, which similarly identifies object couplings through time-windowed mutual information\cite{merlo2025exploiting, rofer2026sparta}.

\textit{Virtual Frame Supplementation} provides additional spatial reference.
Consider \figref{fig:kinematic_analysis}; when using static frames, the mug's frame drives a post-grasp lift.
In dynamic scenes, however, our kinematic filtering removes this newly linked frame to avoid model collapse.
To compensate, we record the end-effector's pose at the start of each skill and introduce it as a static, virtual frame.
(While virtual object frames could serve the same purpose, we opt for end-effector frames as they are easier to track in practice.)
This virtual frame not only drives post-grasp behavior (\figref{fig:virtual_frames}) but also acts as a primitive spatial memory, enabling the policy to, for instance, return a manipulated object to its original location.

\textit{Task-Parameter Selection:} We evaluate the remaining dynamic object frames alongside the newly added virtual frames based on their relative precision~\cite{von2024art}.
For \(C\) candidate frames across \(T\) time steps, we compute
\begin{equation}\label{eq:frame_sel}
    \omega(f) = \max_{t=1}^T \frac{\det \left(\boldsymbol\Sigma_t^{(f)}\right)^{-1}}{\sum_{c=1}^C \det \left(\boldsymbol\Sigma_t^{(c)}\right)^{-1}}
\end{equation}
and select all frames \(f\) over some threshold \(\omega(f)>\tau_{\omega}\).
In \figref{fig:virtual_frames}, the selected frames are highlighted in green. 
Finally, we sequence these individual skill policies to execute the complete long-horizon task~\cite{rozo2020learning, von2024art, von2025unreasonable}.
\algref{alg:dynamic_multistream} summarizes the full procedure and \figref{fig:frame_selection} shows the final frame selection for the dynamic \texttt{PlaceCups} task from \figref{fig:kinematic_analysis}.

\begin{algorithm}[t]
\caption{\ourmethod: Dynamic Multi-Stream Learning}
\label{alg:dynamic_multistream}
\begin{algorithmic}[1]
\Require Initial frames $\{f\}_{f=1}^F$, Thresholds $\tau_M, \tau_{\omega}$
\State $\{t_k\}_{k=1}^K \leftarrow$ Segment task into skills, return start indices
\State $V \leftarrow \emptyset$ \hfill \textcolor{gray}{\(\triangleright\) \textit{Initialize virtual frame history}}

\For{$k = 1, \dots, K$} \hfill \textcolor{gray}{\(\triangleright\) \textit{Per skill}}
    \State $C \leftarrow \left\{ c \in \{f\}_{f=1}^F \mid M_t^{(c)} \ge \tau_M \text{ via \eqref{eq:gmsd}} \right\}$ \hfill \textcolor{gray}{\(\triangleright\) \textit{Links}}
    \State $V \leftarrow V \cup \{\mathit{EE}_{t_k}\}$ \hfill \textcolor{gray}{\(\triangleright\) \textit{Add virtual frames}}
    \State $S \leftarrow \left\{ c \in C \cup V \mid \omega(c) > \tau_{\omega} \text{ via \eqref{eq:frame_sel}} \right\}$ \hfill \textcolor{gray}{\(\triangleright\) \textit{Select}}
    \State Fit multi-stream skill policy on $S$
\EndFor
\State Sequence skill policies
\end{algorithmic}
\end{algorithm}

\subsection{Bimanual Manipulation as Multi-Agent Cooperation}\label{sec:appr_biman}
\begin{figure}[t]
    \vspace{-0.3cm}
    \centering
    \raisebox{.29in}{%
      \includegraphics[scale=.25,
        trim={0 0 0 1.25cm},
        clip]{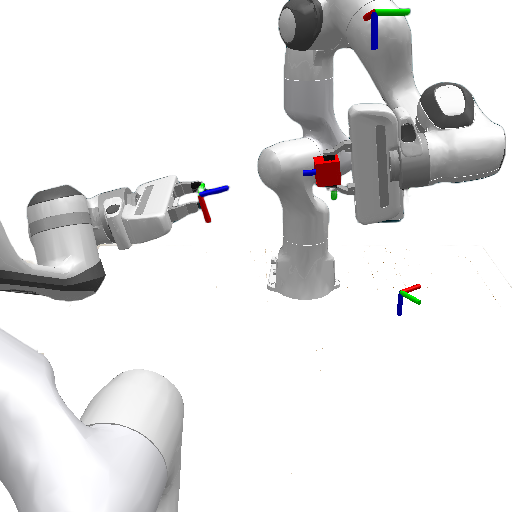}%
    }%
    \includegraphics[height=1.6in]{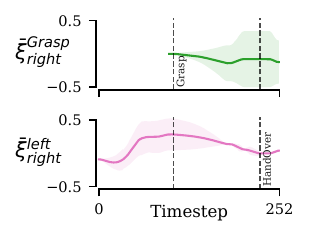}
    \vspace{-0.3cm}
    \caption{Bimanual coordination using dynamic frames.
    Here, the left end-effector \(\pose_{\mathrm{left}}\) guides the right end-effector \(\pose_{\mathrm{right}}\) from grasp to hand-over.}
    \label{fig:biman_frames}
    \vspace{-0.4cm}
\end{figure}

\begin{figure*}[tb]
    \begin{tikzpicture}
        \def\imgwidth{0.125\textwidth}
        \def\imgheight{0.125\textwidth}
        \def\borderoffset{1pt}
        \def\borderwidth{2pt}
         \node[anchor=north west,inner sep=0pt] (leftbox) {
            \includegraphics[width=\imgwidth]{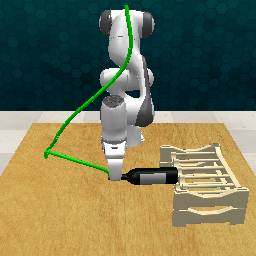}%
            \includegraphics[width=\imgwidth]{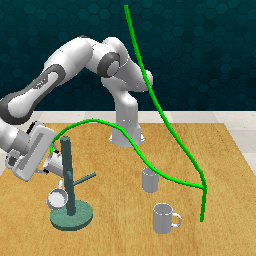}%
            \includegraphics[width=\imgwidth]{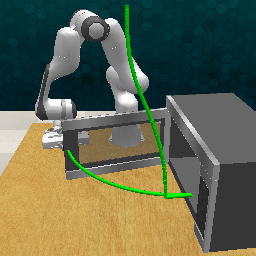}%
            \includegraphics[width=\imgwidth]{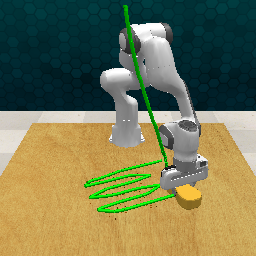}%
            \includegraphics[width=\imgwidth, height=\imgwidth]{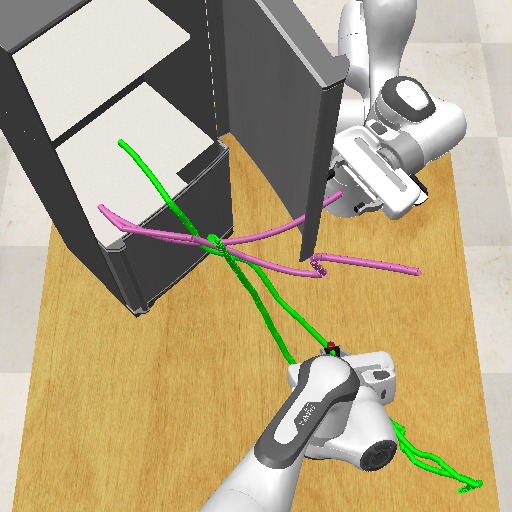}%
            \includegraphics[width=\imgwidth, height=\imgwidth]{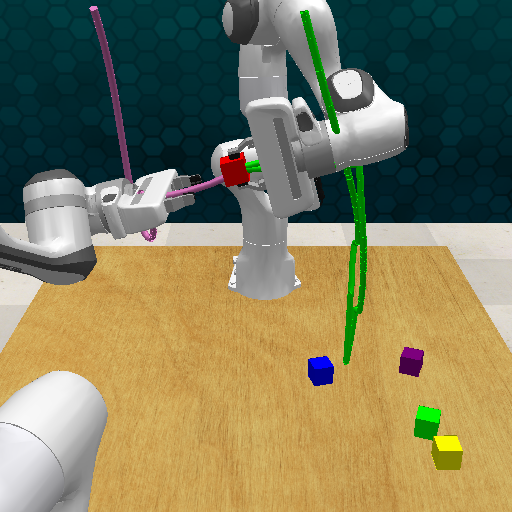}%
            \includegraphics[width=\imgwidth, height=\imgwidth]{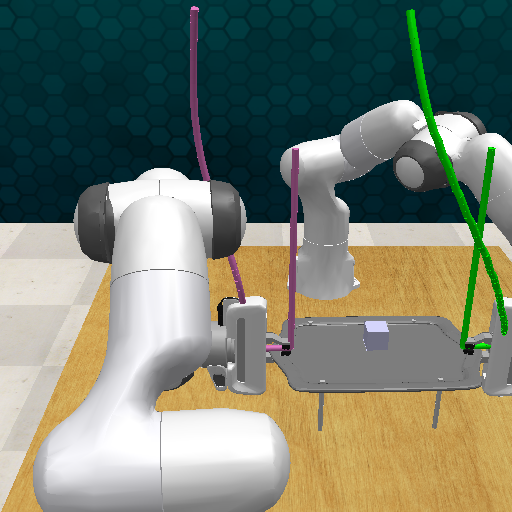}%
            \includegraphics[width=\imgwidth, height=\imgwidth]{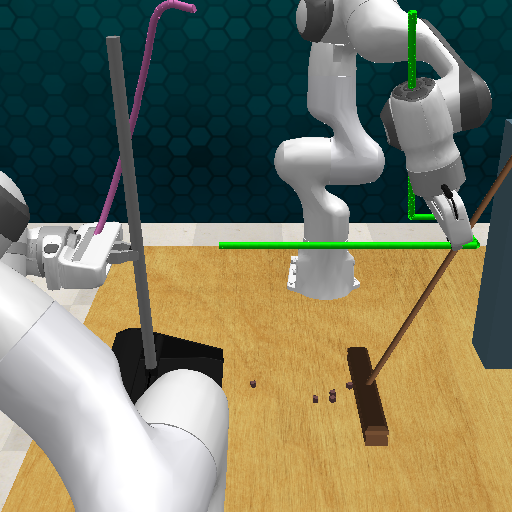}%
            };
        \draw[tabblue, line width=2pt] 
            (0, 0) -- (4*\imgwidth-\borderoffset, 0)  -- (4*\imgwidth-\borderoffset, -\imgheight) -- (0, -\imgheight) -- (0,0);
        \draw[tabred, line width=2pt] 
            (4*\imgwidth+\borderoffset, 0) -- (8*\imgwidth, 0)  -- (8*\imgwidth, -\imgheight) -- (4*\imgwidth+\borderoffset, -\imgheight) -- (4*\imgwidth+\borderoffset,0);
    \end{tikzpicture}
    \caption{RLBench tasks.
    \textcolor{tabblue}{Blue}: The \textit{unimanual} tasks  \texttt{StackWine}, \texttt{PlaceCups}, \texttt{OpenMicrowave}, and \texttt{WipeDesk}.
    \textcolor{tabred}{Red}: The  \emph{bimanual} tasks \texttt{StoreBottle}. \texttt{HandOver}, \texttt{LiftTray} and \texttt{SweepDust}.
    \textcolor{tabpink}{Pink} and \textcolor{tabgreen}{green} trajectories are predicted by \ourmethod{}.
    Object poses are randomized between task instances.
    }\label{fig:sim_tasks}
\end{figure*}

Rather than employing a single monolithic policy or an explicit coordination module, we formulate bimanual manipulation as a multi-agent coordination problem where each arm operates independently.
From the perspective of one arm, the opposite arm acts as another dynamic element in the environment.
This perspective allows us to reduce bimanual coordination directly to two concurrent instances of dynamic multi-stream learning.
By treating the opposite arm's end-effector as a candidate task-parameter, we eliminate the need for hard-coded leader-follower relationships.
Pre-defining a leader fails, for instance, when the dominant arm is not known a priori, or when the roles change dynamically mid-task.
Instead, in our framework, temporary leader roles emerge inevitably from the task-parameter selection: the opposite end-effector simply acts as a dominant spatial focal point whenever appropriate.
To implement this, we fit a policy for each arm using \algref{alg:dynamic_multistream}.
We augment the initial candidate set \(C=\{\pose_f\}_{f=1}^F\) with the dynamic frame of the opposite end-effector (i.e., adding \(\pose_\mathrm{left}\) to the right arm's candidate set, and vice versa).
To see why, contrast the \texttt{LiftTray} and \texttt{HandOver} tasks in \figref{fig:sim_tasks}.
In \texttt{LiftTray}, the tray serves as a shared spatial reference, meaning the arms only need to coordinate temporally.
In contrast, as the handover lacks a static environmental reference, the arms must dynamically coordinate a rendezvous point.
Whether the demonstrated strategy relied on a dominant leader arm to establish this point or a mutual convergence, the underlying causal structure is captured by the dynamic policy streams (see \secref{sec:man_dyn_env} and \figref{fig:biman_frames}).
Just as \(p(\pose_{\ee}\,|\,\mathrm{Mug})\) modeled the mug's causal influence in \figref{fig:kinematic_analysis}, the stream \(p(\pose_{\mathrm{right}}\,|\,\pose_{\mathrm{left}})\) models the left arm's influence on the right in \figref{fig:biman_frames}.
If the spatial convergence was mutual, this bilateral influence is reflected across both sets of policy streams.
Thus, solving two concurrent instances of \ourmethod{} resolves the bimanual coordination problem.

During a handover, the object pose \(\pose_\mathrm{cube}\) and the grasping arm's pose \(\pose_\mathrm{left}\) carry redundant  information for the receiving arm \(\pose_\mathrm{right}\).
However, visual object tracking degrades under occlusions.
Leveraging the opposite arm's precise forward kinematics as a task parameter circumvents this vulnerability (\secref{sec:exp_real}).
Furthermore, per our kinematic analysis (\secref{sec:man_dyn_env}), \ourmethod{} is applicable even if the two robots are kinematically linked, which might be useful in mobile manipulation.

\begin{table*}
        \caption{Policy success rates on unimanual RLBench tasks. }\label{tab:success_rates_uni}
        \centering
        \setlength{\tabcolsep}{4.5pt}
        \begin{threeparttable}
        \begin{tabular}{ll ccccc cccc cccc c}
            \toprule
            \multirow{2}{*}{\textbf{Demos}} & \multirow{2}{*}{\textbf{Method}} & \multicolumn{5}{c}{\textbf{Static Environment}} & \multicolumn{4}{c}{\textbf{Zero Shot: Smooth Dynamics}} & \multicolumn{4}{c}{\textbf{Zero Shot: Teleportation}} & \multirow{2}{*}{\textbf{Avg.}} \\
            \cmidrule(lr){3-7} \cmidrule(lr){8-11}\cmidrule{12-15}
            & & \makecell{Stack \\ Wine} & \makecell{Place \\ Cups} & \makecell{Open \\ Microw.} & \makecell{Wipe \\ Desk} & Avg. & \makecell{Stack \\ Wine} & \makecell{Place \\ Cups} & \makecell{Open \\ Microw.} & \makecell{Wipe \\ Desk} & \makecell{Stack \\ Wine} & \makecell{Place \\ Cups} & \makecell{Open \\ Microw.} & \makecell{Wipe \\ Desk} & \\
            \midrule
            100 & Diffusion Policy~\cite{chi2023diffusionpolicy} & 0.95 & 0.40 & 0.92 & 0.06 & 0.58 & 0.96 & 0.39 & 0.98 & 0.09 & 0.96 & 0.35 & 0.95 & 0.08 & 0.59\\
            \cmidrule{1-16}
            5 & Diffusion Policy~\cite{chi2023diffusionpolicy} & 0.18 & 0.01 & 0.01 & 0.00 & 0.05 & 0.16 & 0.00 & 0.01 & 0.00 & 0.19 &0.00 & 0.01 & 0.00 & 0.05 \\
            & ARP + RRTC\textsuperscript{*}~\cite{zhang2024arp} & 0.64 & 0.02 & 0.07 & 0.00 & 0.18 & 0.00 & 0.00 & 0.00 & 0.00 & 0.00 & 0.00 & 0.00 & 0.00 & 0.06 \\
            & TAPAS-GMM\textsuperscript{\textdagger}~\cite{von2024art} & \textbf{1.00} & 0.93 & 0.00 & 0.01 & 0.49 & 0.00 & 0.00 & 0.00 & 0.00 & 0.00 & 0.00 & 0.00 & 0.00 & 0.16 \\
            & MiDiGaP\textsuperscript{\textdagger}~\cite{von2025unreasonable} & \textbf{1.00} & 0.97 & \textbf{0.99} & \textbf{1.00} & 0.99 & 0.00 & 0.00 & 0.00 & 0.00 & 0.00 & 0.00 & 0.00 & 0.00 & 0.32\\
            & \textbf{\ourmethod{} (Ours)} & \textbf{1.00} & \textbf{0.99} & \textbf{0.99} & \textbf{1.00} & \textbf{1.00} & \textbf{1.00} & \textbf{0.97} & \textbf{0.99} & \textbf{0.66} & \textbf{1.00} & \textbf{0.99} & \textbf{0.97} & \textbf{0.69} & \textbf{0.94} \\
            \bottomrule
        \end{tabular}
        \begin{tablenotes}[para,flushleft]
           \footnotesize  
           Bold values indicate the best-performing model per task.
           {*} indicates that ARP predicts keyposes only.
           {\textdagger} denotes static-frame policies.
         \end{tablenotes}
        \end{threeparttable}
        \vspace{-.4cm}
\end{table*}

\section{\ourbench{}: A Dynamic Environment Manipulation Benchmark}\label{sec:bench}
We build \ourbench{} on RLBench~\cite{james2019rlbench}, leveraging its automated demonstration generation and repository of over 100 diverse tasks.
RLBench has spawned numerous extensions, including bimanual manipulation~\cite{grotz2024peract2}, visual perturbation~\cite{pumacay2024colosseum}, task composition~\cite{zheng2022vlmbench}, and memory~\cite{huang2026robostream}.
By integrating with this ecosystem, \ourbench{} enables researchers to jointly evaluate these distinct axes of generalization while preserving the underlying data generation and task diversity.

A key challenge in defining a robotic manipulation benchmark is task instance variability.
To rigorously evaluate the generalization capabilities of a policy, the object placement should vary as widely as possible across task instances, while remaining kinematically feasible.
While many benchmarks restrict objects to a conservative safe spawn zone, RLBench employs a sample-then-verify scheme: it samples objects across a wide workspace, verifies kinematic feasibility, and resamples failures.
This approach maximizes task instance diversity~\cite{von2025unreasonable}.

This tension between variability and feasibility intensifies in dynamic scenes, where moving objects must remain reachable over time.
We resolve this by independently sampling and verifying two separate, valid static task configurations.
The environment initializes at the first configuration and interpolates toward the second, either via smooth spatial interpolation or abrupt teleportation (\panelref{subfig:move} in \figref{fig:kinematic_analysis}).
Since both boundary configurations are independently validated, this strategy guarantees kinematic feasibility while maximizing initial configuration diversity.
Furthermore, interpolating between arbitrary task configurations automatically varies the direction, distance, and velocity of object movements across episodes.
Multiple such environment transformations can be chained, and their timing and speed can be controlled.
We predefine a set of sensible defaults based on average task duration.
Qualitative examples are shown in the supplementary video.

\section{Experimental Evaluation}\label{sec:exp}
We evaluate \ourmethod{} in four parts.
\secref{sec:exp_uni} examines unimanual tasks. Is \ourmethod{} competitive in static environments? Does it transfer to dynamic environments, adapting to both smooth and abrupt object dynamics?
\secref{sec:exp_bi} turns to bimanual tasks. Does \ourmethod{} enable effective bimanual coordination?
How does it perform relative to other approaches, such as leader-follower methods and monolithic policies?
\secref{sec:exp_bi_dyna} then investigates dynamic coordination.
Does it dynamically re-coordinate if an object or an arm is perturbed during inference?
Finally, \secref{sec:exp_real} verifies our simulation results on a real, bimanual robot.
The supplementary video analyses qualitative examples, including failure cases, of our experiments.

\subsection{Unimanual Static and Dynamic Manipulation}\label{sec:exp_uni}
\para{Tasks: }
We evaluate the unimanual policy learning capabilities of \ourmethod{} on a representative set of RLBench tasks used in prior work~\cite{von2025unreasonable}.
The tasks, shown in \figref{fig:sim_tasks}, include pick and place (\texttt{StackWine}), high-precision (\texttt{PlaceCups}), articulated object interaction (\texttt{OpenMicrowave}) and highly-constrained movements (\texttt{WipeDesk}).
We evaluate three variants of each task: the standard static environment setup, as well as two variants of zero-shot policy transfer on \ourbench.
\textit{Smooth dynamics} moves the object poses over a short time horizon, whereas \textit{teleportation} moves them abruptly.

\para{Baselines:} 
We compare against a set of deep learning methods, including the Diffusion Policy~\cite{chi2023diffusionpolicy} and the Auto-Regressive Policy (ARP)~\cite{zhang2024arp}, the current incumbent transformer-based policy architecture.
ARP only predicts the next keypose for the end-effector, not a dense trajectory, and leverages the RRT-Connect planner~\cite{kuffner2000rrtc} with access to ground-truth collision information.
We further compare against TAPAS-GMM~\cite{von2024art} and MiDiGaP~\cite{von2025unreasonable}, two state-of-the-art multi-stream learning methods, trained on five demonstrations.
To disentangle policy learning and representation learning, all methods receive ground-truth object pose observations.
We evaluate learning from RGB-D observations in \secref{sec:exp_real}.
ARP \emph{additionally} relies on four RGB-D cameras for 3D scene reconstruction.
We report task success rates over 200 evaluation episodes in \tabref{tab:success_rates_uni}.

\para{Results: }
\tabref{tab:success_rates_uni} paints a sharp contrast.
Vanilla MiDiGaP achieves robust performance in the static environments, comfortably outperforming the other baselines.
However, both TAPAS-GMM and MiDiGaP fail completely in the dynamic environments, as the exogeneity of the task parameters is violated.
ARP similarly lacks a mechanism to react to the environment dynamics.
Diffusion Policy generalizes well to dynamic environments, but is hampered by its sample inefficiency.
In contrast, \ourmethod{} achieves substantially higher performance, outperforming Diffusion Policy by 35 percentage points while using 20-times less data.
At data parity, this advantage grows to 89 percentage points.
\ourmethod{} not only achieves top performance in the static environments, but also robustly generalizes to the dynamic environments as well.
This \textbf{zero-shot} capability is an exciting advantage of \ourmethod{}.
Collecting demonstrations is significantly easier in static environments than in dynamic ones.
By generalizing zero-shot, \ourmethod{} drastically reduces the effort required to train manipulation skills for dynamic environments.

\para{Ablation Study:} Omitting \ourmethod{}'s kinematic analysis leads to universal task failure due to causal collapse (\secref{sec:man_dyn_env}).
Omitting the virtual frame supplementation removes spatial reference, primarily causing jerkier trajectories on \texttt{PlaceCups} and \texttt{StackWine}, but near-total task failure on \texttt{WipeDesk} and \texttt{OpenMicrowave}.
Task-parameter selection does not significantly influence policy performance due to the precision weighting of the streams, but it speeds up both learning and inference by excluding irrelevant frames.

\subsection{Static Bimanual Manipulation}\label{sec:exp_bi}
We adopt four representative tasks from RLBench2~\cite{grotz2024peract2}, shown in \figref{fig:sim_tasks}.
These tasks encompass long horizons and interaction with articulated objects (\texttt{StoreBottle}), precise temporal, spatial, and physical coordination (\texttt{HandOver}), as well as both symmetric (\texttt{LiftTray}) and asymmetric cooperation (\texttt{SweepDust}).
We compare against a set of monolithic and leader-follower transformer-based architectures~\cite{grotz2024peract2, zhao2023learning}, as well as monolithic Diffusion Policy and MiDiGaP.

\begin{table}
        \caption{Policy success rates on static bimanual RLBench tasks. }\label{tab:success_rates_bi}
        \centering
        \setlength{\tabcolsep}{5pt}
        \begin{threeparttable}
        \begin{tabular}{ll cccc c}
            \toprule
            \textbf{Demos}& \textbf{Method} & \makecell{Store \\ Bottle} & \makecell{Hand \\ Over} & \makecell{Sweep \\ Dust} & \makecell{Lift \\ Tray} &  Avg. \\
            \midrule
            100 & Diffusion Policy~\cite{chi2023diffusionpolicy} & 0.01 & 0.48 & 0.98 & 0.82 & 0.57 \\
            & ACT\textsuperscript{*}~\cite{zhao2023learning} & 0.00 & 0.00 & 0.00 & 0.06 & 0.02 \\
            & RVT-LF\textsuperscript{*}~\cite{grotz2024peract2} & 0.00 & 0.00 & 0.00 & 0.06 & 0.02 \\
            & PerAct-LF\textsuperscript{*}~\cite{grotz2024peract2} & 0.00 & 0.00 & 0.28 & 0.14 & 0.11\\
            & PerAct2\textsuperscript{*}~\cite{grotz2024peract2} & 0.03 & 0.11 & 0.00 & 0.01 & 0.04 \\ 
            \cmidrule{1-7}
            5 & Diffusion Policy~\cite{chi2023diffusionpolicy} & 0.00 & 0.01 & 0.78 & 0.22 & 0.25 \\
            & MiDiGaP~\cite{von2025unreasonable} & \textbf{0.82} & \textbf{0.97} &  \textbf{1.00} & \textbf{1.00} & \textbf{0.95} \\
            & \textbf{\ourmethod{} (Ours)} & \textbf{0.82} & \textbf{0.97} &  \textbf{1.00} & \textbf{1.00} & \textbf{0.95} \\
            \bottomrule
        \end{tabular}
        \begin{tablenotes}[para,flushleft]
           \footnotesize  
           {*} denotes results from Grotz \etal{}\cite{grotz2024peract2}. LF denotes leader--follower.
         \end{tablenotes}
        \end{threeparttable}
\end{table}

\begin{table}
        \caption{Policy success rates on dynamic bimanual RLBench tasks. }\label{tab:success_rates_bi_dyn}
        \centering
        \setlength{\tabcolsep}{2.75pt}
        \begin{threeparttable}
        \begin{tabular}{ll cc cccc c}
            \toprule
            \multirow{2}{*}{\textbf{Demos}} & \multirow{2}{*}{\textbf{Method}} & \multicolumn{2}{c}{\textbf{Coordination}} & \multicolumn{4}{c}{\textbf{Dynamic Environment}} & \multirow{2}{*}{\textbf{Avg.}}\\
            \cmidrule(lr){3-4}\cmidrule(lr){5-8}
            & & \makecell{Hand \\ Left} & \makecell{Hand \\ Right} &  \makecell{Store \\ Bottle} & \makecell{Hand \\ Over} & \makecell{Sweep \\ Dust} & \makecell{Lift \\ Tray} & \\
            \midrule
            100 & Diffusion~\cite{chi2023diffusionpolicy} & 0.01 & 0.04 & 0.00 & 0.43 & 0.98 & 0.67 & 0.36 \\
            \cmidrule{1-9}
                5 & Diffusion~\cite{chi2023diffusionpolicy} &  0.00 & 0.00 & 0.00 & 0.00 & 0.79 & 0.24 & 0.17 \\
            & MiDiGaP\cite{von2025unreasonable} &  0.42 & 0.01 & 0.05 & 0.04 & \textbf{1.00} & 0.05 & 0.26 \\
            & \textbf{\ourmethod{}} & \textbf{0.97} & \textbf{0.97} & \textbf{0.82} & \textbf{0.97} &  \textbf{1.00} & \textbf{1.00} & \textbf{0.96} \\
            \bottomrule
        \end{tabular}
        \end{threeparttable}
        \vspace{-0.3cm}
\end{table}

\tabref{tab:success_rates_bi} shows that both monolithic and leader-follower transformer policies struggle to time and coordinate arm movements.
They grasp imprecisely and deviate from the required trajectories.
The Diffusion Policy succeeds on short-horizon, low-variability tasks (\texttt{SweepDust}), but fails as task complexity increases (\texttt{StoreBottle}) - even when given 100 demonstrations.
\ourmethod{} robustly outperforms all baselines, save MiDiGaP, while requiring significantly fewer training samples.
Given 20-times less data, it outperforms Diffusion Policy by 38 percentage points; at data parity, the margin increases to 70 percentage points.
MiDiGaP performs on par with \ourmethod{} on these four static bimanual tasks because they offer a clear focal point for coordination, for instance, the tray's pose for joint lifting (see \secref{sec:appr_biman}).

\begin{figure}[t]
    \centering
    \def\imgwidth{0.25\linewidth}
    \def\imgheight{0.25\linewidth}
    \includegraphics[width=\imgwidth, height=\imgheight]{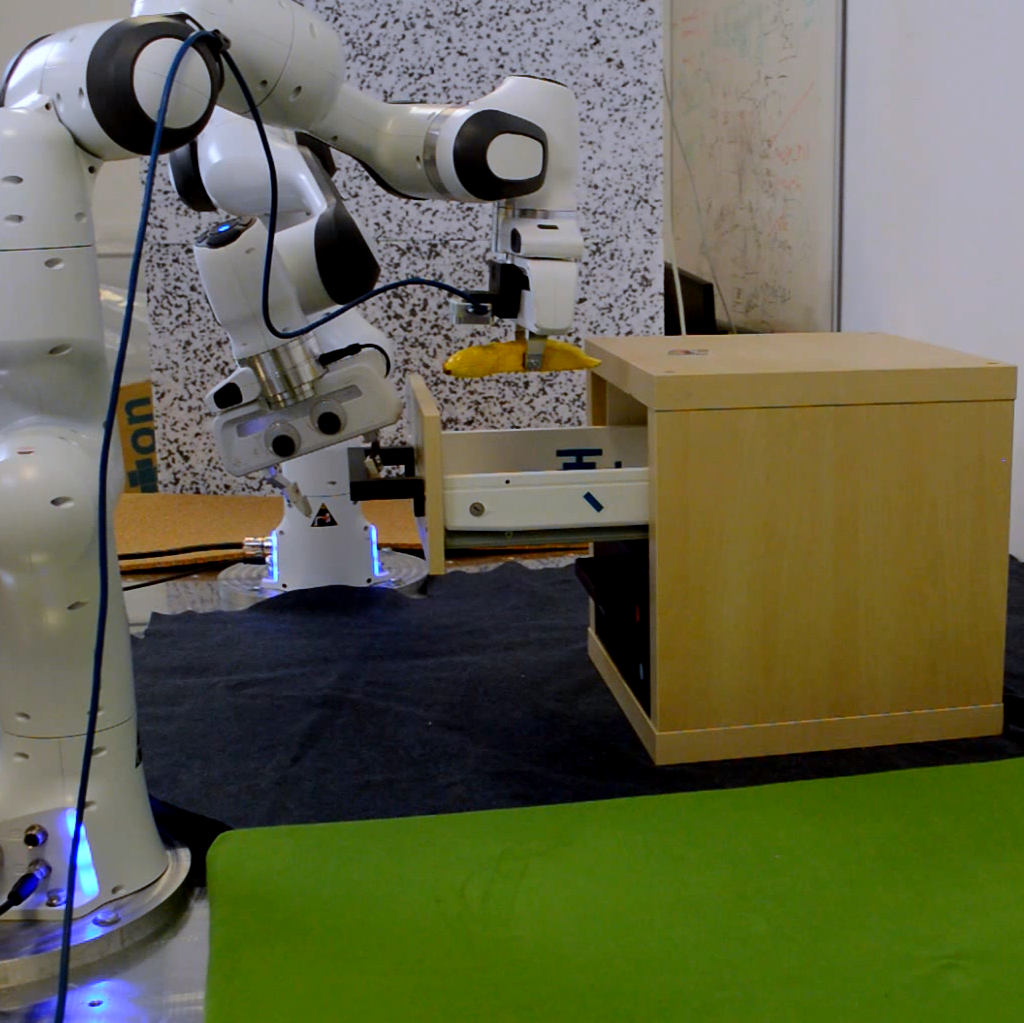}%
    \includegraphics[width=\imgwidth, height=\imgheight]{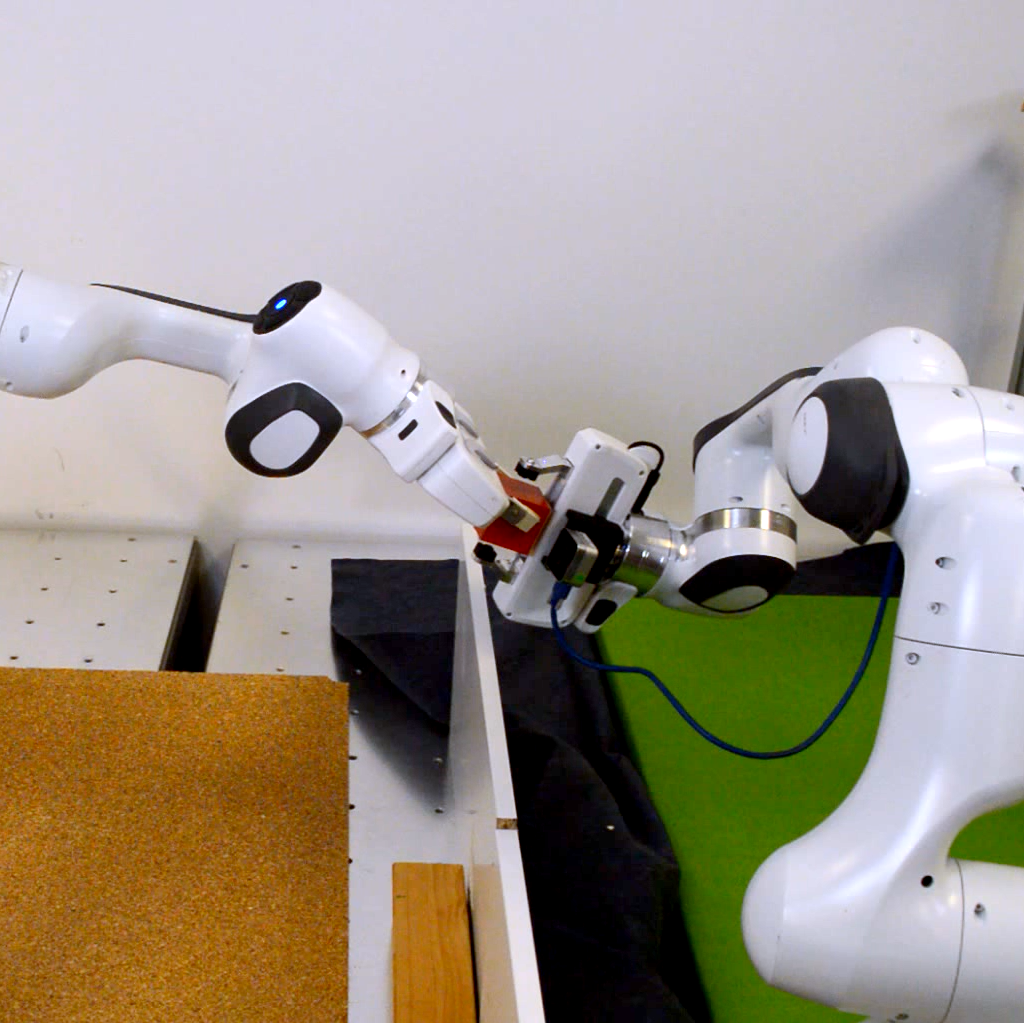}%
    \includegraphics[width=\imgwidth, height=\imgheight]{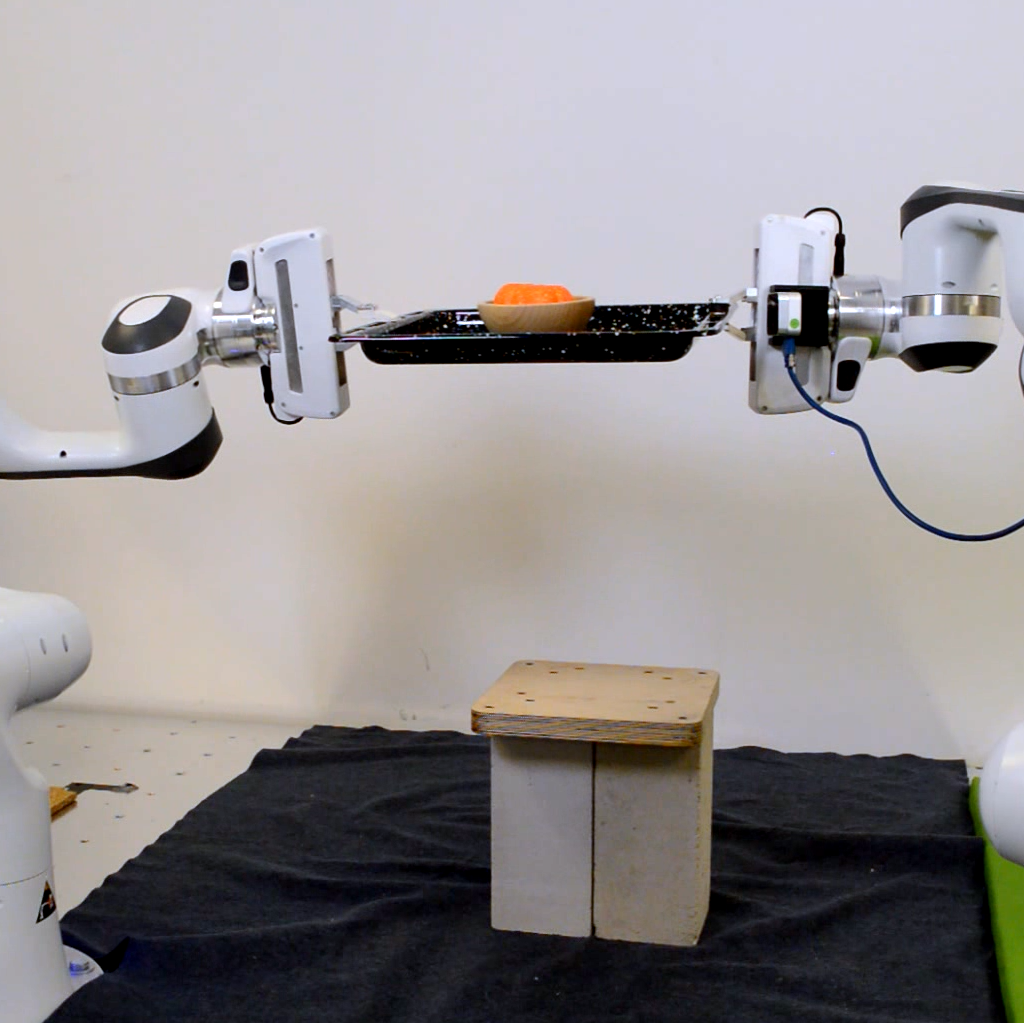}%
    \includegraphics[width=\imgwidth, height=\imgheight]{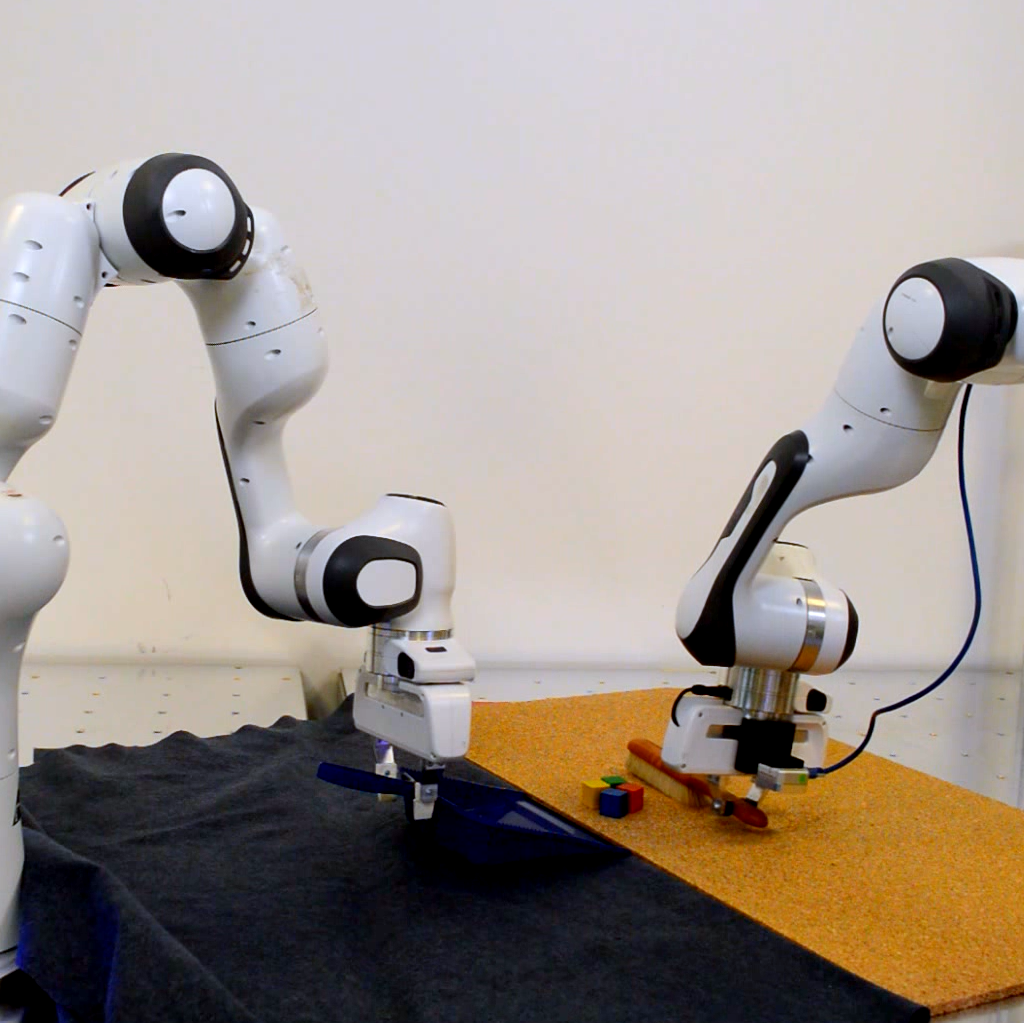}%
    \caption{Real-world tasks: \texttt{StoreItem}, \texttt{HandOver}, \texttt{LiftJointly}, and \texttt{SweepBlocks}.
    Object poses are randomized between task instances.
    }\label{fig:real_tasks}
\end{figure}%
\begin{table}[t]
        \caption{Policy success rates on bimanual real-world tasks. }\label{tab:success_rates_real}
        \centering
        \setlength{\tabcolsep}{2pt}
        \begin{threeparttable}
        \begin{tabular}{l cccc cccc c}
            \toprule
            \multirow{2}{*}{\textbf{Method}} & \multicolumn{4}{c}{\textbf{Static Environment}} & \multicolumn{2}{c}{\textbf{Dynamic}} & \multicolumn{2}{c}{\textbf{Coordination}} & \multirow{2}{*}{\textbf{Avg.}} \\
            \cmidrule(lr){2-5}\cmidrule(lr){6-7}\cmidrule(lr){8-9}
             & \makecell{Store \\ Item} & \makecell{Hand \\ Over} & \makecell{Sweep \\ Blocks} & \makecell{Lift \\ Jointly} & \makecell{Store \\ Item} & \makecell{Hand \\ Over} & \makecell{Hand \\ Left} & \makecell{Hand \\ Right}\\
            \midrule
            Diffusion~\cite{chi2023diffusionpolicy} & 0.04 & 0.00 & 0.08 & 0.04 & 0.00 & 0.00 & 0.00 & 0.00 & 0.02 \\
            MiDiGaP~\cite{von2025unreasonable} & 0.96 & 0.32 & 0.96 & 1.00 & 0.00 & 0.00 & 0.00 & 0.00 & 0.41 \\
            \textbf{\ourmethod{}} & 0.96 & 0.96 & 0.96 & 1.00 & 0.92 & 0.88 & 0.92 & 0.96 & 0.95 \\
            \bottomrule
        \end{tabular}
        \begin{tablenotes}[para,flushleft]
           \footnotesize  
           Dynamic: perturbation of task objects. Coordination: perturbation of robot.
         \end{tablenotes}
        \end{threeparttable}
        \vspace{-0.3cm}
\end{table}

\subsection{Dynamic Bimanual Coordination}\label{sec:exp_bi_dyna}
While RLBench2 offers challenging manipulation tasks, the bimanual behavior is pre-coordinated, and the environments remain static.
For instance, in \texttt{HandOver}, the handover location is pre-defined relative to the task space.
Consequently, the two arms do not need to spatially coordinate with each other and can instead orient themselves independently.
Furthermore, task objects are never manipulated or disturbed by external agents.
To evaluate \emph{dynamic} coordination and dynamic environments, we integrate \ourbench{} with RLBench2 and perform two sets of experiments.
First, we introduce a modified version of \texttt{HandOver} where the handover location is randomized.
During inference, we then perturb the predicted trajectory of one arm to test whether the other arm adaptively responds.
Second, we apply the dynamic task augmentation, described in \secref{sec:bench}, to all four bimanual tasks.

While MiDiGaP performs strongly on the static tasks (\tabref{tab:success_rates_bi}), it fails as soon as dynamic coordination or reactivity to environment dynamics is required (\tabref{tab:success_rates_bi_dyn}).
The only outlier is \texttt{SweepDust}. While the dust positions are randomized across task instances, their dispersion is so low that different instances can be solved with the same movement.
Diffusion Policy often reacts effectively to environment dynamics, only incurring modest performance penalties.
However, it falters completely when one of the arms is perturbed, failing to dynamically re-coordinate.
In contrast, \ourmethod{} remains resilient against both challenges, demonstrating robust performance across all scenarios.
Besides reacting effectively to environment dynamics, it re-coordinates dynamically if either arm is perturbed. 
\ourmethod{}'s zero-shot capability to dynamically adapt to perturbations of one arm opens promising avenues for future research, particularly in human–robot collaboration. 
Unlike a robot, a human collaborator may not perform a task in a fully consistent manner but may vary their behavior across trials.
Due to the observed robustness to perturbations, \ourmethod{} may generalize to settings involving variable human behavior.
We perform a first foray in the next section.

\subsection{Real World Experiments}\label{sec:exp_real}
We verify our simulated experiments on a real, bimanual Franka Emika robot with a single RealSense D405 RGB-D camera.
The tasks, shown in \figref{fig:real_tasks}, recreate the challenges of the simulated experiments, including object storage, articulated object interaction, handover, joint lifting, and asymmetric coordination.
For each task, we collect five demonstrations.
As in simulation, we evaluate both static and zero-shot dynamic conditions comprising object dynamics and arm perturbations, each for 25 episodes per scenario.
Following prior work, we condition all policies on task parameters estimated from DINO features of the RGB-D observation~\cite{von2024art, von2025unreasonable}.

\tabref{tab:success_rates_real} confirms our simulated results.
MiDiGaP performs well in the static scenarios and when there is a clear focal point for coordination, such as an object to be lifted jointly by both arms.
However, it fails to adapt to environment dynamics, and when no common focal point is present.
For example, it struggles to coordinate free-space hand-over given inconsistent hand-over locations in the demonstrations.
Diffusion Policy is not sufficiently sample-efficient to learn any of the challenging real-world tasks.
In contrast, \ourmethod{} performs on par with MiDiGaP in the static scenarios, and generalizes effectively to the dynamic variants.
By not relying on a static focal point, it excels even in free-space handover, as well as competently re-coordinating dynamically if either a task object or one of the arms is perturbed.
Most remaining failure cases are due to perception errors - most often faulty depth measurements.

To evaluate \ourmethod{} for human-robot collaboration, we modify the \texttt{HandOver} task by withholding the opposite arm's end-effector pose, forcing the policy to rely exclusively on the tracked object.
Under these conditions, success hinges heavily on the camera viewpoint.
As long as the object remains visible, the policy performs competitively.
Under forced occlusion, however, success rates plummet, a vulnerability that more robust object tracking could mitigate~\cite{vonhartz2023treachery}.
Crucially, when we replace the transmitting robotic arm with a human, the policy generalizes zero-shot to human-robot handovers. 

\section{Conclusion}
We presented \ourmethod{}, a framework that unifies bimanual manipulation, multi-agent cooperation, and interaction with dynamic environments under the umbrella of dynamic multi-stream learning.
As a lightweight, policy-agnostic layer, it preserves the native of multi-stream policies, such as outstanding sample efficiency.
Through extensive experiments, we demonstrated that \ourmethod{} effectively reacts to environment dynamics while coordinating multiple arms without requiring an explicit hierarchy or dedicated coordination module. 
Crucially, \ourmethod{} operates in dynamic environments through zero-shot generalization from static demonstrations, substantially simplifying data collection.
Furthermore, it generalizes zero-shot to perturbations affecting either arm, demonstrating robust adaptation to previously unseen interaction dynamics.

\para{Limitations and Future Work:}
By building on multi-stream learning, \ourmethod{} inherits some of its structural limitations, namely the requirement that individual skills can be segmented from long-horizon tasks~\cite{von2024art}, and the reliance on an external vision encoder module.
However, this modularity is also a strength, as it enables harnessing novel computer vision methods without policy retraining.
Future work might investigate \ourmethod{}'s potential for human-robot collaboration and whole-body coordination in mobile manipulation.

\bibliographystyle{IEEEtran}
\bibliography{IEEEabrv,root}

\end{document}